\documentclass{article}

     \usepackage[nonatbib,preprint]{neurips_2019}

\usepackage[utf8]{inputenc} 
\usepackage[T1]{fontenc}   
\usepackage{hyperref}
\usepackage{url}
\usepackage{amsmath}
\usepackage[pdftex]{graphicx}
\usepackage{subfig}
\usepackage{tikz}
\usepackage{multirow}
\usepackage{nicefrac}
\usepackage{booktabs}
\usepackage{xcolor}
\usepackage{colortbl}
\usepackage{float}
\usepackage{microtype}

\newcommand{\bfu}{\mathbf{u}}
\newcommand{\bfv}{\mathbf{v}}

\newcommand{\bfx}{\mathbf{x}}
\newcommand{\bfy}{\mathbf{y}}
\newcommand{\bfz}{\mathbf{z}}

\newcommand{\bfI}{\mathbf{I}}
\newcommand{\bfJ}{\mathbf{J}}
\newcommand{\bfU}{\mathbf{U}}
\newcommand{\bfV}{\mathbf{V}}

\newcommand{\bfSigma}{\boldsymbol{\Sigma}}

\newcommand{\bfgamma}{\boldsymbol{\gamma}}
\newcommand{\bfbeta}{\boldsymbol{\beta}}
\newcommand{\bftheta}{\boldsymbol{\theta}}
\newcommand{\bfmu}{\boldsymbol{\mu}}
\newcommand{\bfnu}{\boldsymbol{\nu}}
\newcommand{\bfeta}{\boldsymbol{\eta}}

\newcommand{\bfcalB}{\boldsymbol{\mathcal{B}}}
\newcommand{\bfcalC}{\boldsymbol{\mathcal{C}}}
\newcommand{\bfcalF}{\boldsymbol{\mathcal{F}}}
\newcommand{\bfcalN}{\boldsymbol{\mathcal{N}}}
\newcommand{\bfcalR}{\boldsymbol{\mathcal{R}}}
\newcommand{\bfzero}{\mathbf{0}}

\makeatletter
\newcommand{\printfnsymbol}[1]{%
  \textsuperscript{\@fnsymbol{#1}}%
}
\makeatother

\title{Residual Networks as Nonlinear Systems: Stability Analysis using Linearization}

\author{
 Kai Rothauge \\
  \small ICSI and Department of Statistics \\
  \small UC Berkeley \\
  \small Berkeley, USA \\
  \small \texttt{kai.rothauge@berkeley.edu} \\
   \And
 Zhewei Yao \\
  \small Department of Mathematics \\
  \small UC Berkeley \\
  \small Berkeley, USA \\
  \small \texttt{zheweiy@berkeley.edu} \\
   \AND
 Zixi Hu \\
  \small Department of Mathematics \\
  \small UC Berkeley \\
  \small Berkeley, USA \\
  \small \texttt{zixihu@berkeley.edu} \\
   \And
 Michael W. Mahoney \\
  \small ICSI and Department of Statistics \\
  \small UC Berkeley \\
  \small Berkeley, USA \\
  \small \texttt{mmahoney@stat.berkeley.edu} \\
}

\begin{document}

\maketitle

\begin{abstract}
We regard pre-trained residual networks (ResNets) as nonlinear systems and use linearization, a common method used in the qualitative analysis of nonlinear systems, to understand the behavior of the networks under small perturbations of the input images. We work with ResNet-56 and ResNet-110 trained on the CIFAR-10 data set. We linearize these networks at the level of residual units and network stages, and the singular value decomposition is used in the stability analysis of these components. It is found that most of the singular values of the linearizations of residual units are 1 and, in spite of the fact that the linearizations depend directly on the activation maps, the singular values differ only slightly for different input images. However, adjusting the scaling of the skip connection or the values of the weights in a residual unit has a significant impact on the singular value distributions. Inspection of how random and adversarial perturbations of input images propagate through the network reveals that there is a dramatic jump in the magnitude of adversarial perturbations towards the end of the final stage of the network that is not present in the case of random perturbations. We attempt to gain a better understanding of this phenomenon by projecting the perturbations onto singular vectors of the linearizations of the residual units. 
\end{abstract}

\section{Introduction}

In this work we regard a residual network (ResNet) as nonlinear system
\begin{equation*}
    \bfx_\mathrm{pred} = \bfcalN(\bfx_0; \bftheta).
\end{equation*}
In the context of image classification, the network attempts to classify an input image $\bfx_0$ as belonging to one of $K$ classes, where $\bfx_\mathrm{pred}$ is the output of the network that provides a probability of $\bfx_0$ belonging to each of those classes. The network has a set of parameters $\bftheta$ that is found when training the network, but we only consider pre-trained networks and therefore $\bftheta$ is fixed.

ResNets~\cite{he2016a} have proven to be highly successful in practice, but a deeper understanding why they work as well as they do is still the subject of active research. Much of the research effort is devoted to understanding the behavior of the networks during the training process, but understanding the behavior of pre-trained networks during inference is important as well, particularly with regard to adversarial images that cause otherwise correctly classified images to be misclassified, and serves as the motivation for this work. 

A common approach to studying nonlinear systems is to linearize them~\cite{khalil2001, slotine1991} and then investigate their stability properties, i.e. investigate how their behavior changes if the input is slightly perturbed. Our primary goal is to empirically investigate to what extent linearization can help us understand the behavior of pre-trained ResNets and to determine future directions of exploration. We are motivated by the following questions:
\begin{itemize}
    \item How stable are ResNets? In other words, how do perturbations in the input images grow or shrink as they propagate through the networks?
    \item To what extent do the stability properties depend on the input images? 
    \item Is there significant difference between the behavior of adversarial perturbations compared to random perturbations, and is linearization a useful tool to help in its analysis?
\end{itemize}

\subsection{Linearization}

If $\bfcalF$ is some arbitrary, sufficiently differentiable nonlinear function with input $\bfy$ and output $\bfz$, the perturbation of the output in response to a perturbation $\delta \bfy$ in the input is given by
\begin{equation*}
    \delta \bfz = \bfcalF(\bfy + \delta \bfy) - \bfcalF(\bfy) = \dfrac{d\bfcalF}{d\bfy}\delta \bfy + O(\|\delta \bfy\|^2).
\end{equation*}
The higher-order terms can be neglected if the input perturbation $\delta \bfy$ is sufficiently small and the \textit{linearized system} is then given by
\begin{equation*}
    \delta \bfz \approx \bfJ\,\delta \bfy,
\end{equation*}
where $\bfJ = \frac{d\bfcalF}{ d\,\bfy}$ is the \textit{linearization of $\bfcalF$}, commonly referred to as the \textit{Jacobian} or \textit{sensitivity matrix}.

ResNets can be regarded as the compositions of several network stages, which in turn are compositions of several residual units; we go into more detail in Section 2. To address the above motivating questions, we focus primarily on the linearizations of the residual units, but also briefly look at the linearizations of the network stages.

To study the network's stability properties, we need to inspect the spectra of the linearizations by computing their singular value decomposition (SVD). 

\subsection{Stability Analysis}

The SVD of a $K \times N$ Jacobian is $\bfJ = \bfU\bfSigma\bfV^{*}$, where $\bfU$ is a $K \times K$ real or complex unitary matrix of left-singular vectors $\bfu_i$, $\bfSigma$ is an $K \times N$ rectangular diagonal matrix with the singular values (non-negative real numbers) $\sigma_i$ on the diagonal, and $\bfV$ is an $N \times N$ real or complex unitary matrix of right-singular vectors $\bfv_i$.

If a perturbation $\delta \bfy$ can be written as a linear combination of right-singular vectors of $\bfJ$, $\delta \bfy = \sum\limits_{i=1}\alpha_i\bfv_i$, then, using the fact that the $\bfv_i$ are orthonormal, we have that $\alpha_i = \langle \delta\bfy, \bfv_i \rangle$ and it is easy to show that
\begin{equation*}
\delta \bfz = \bfJ \delta \bfy = \sum\limits_{k=1}\alpha_k\sigma_k\bfu_k,
\end{equation*}
therefore the perturbation of the output is a linear combination of the left-singular vectors of $\bfJ$. The magnitude of $\delta \bfz$ is given by
\begin{equation}
\left\| \delta\bfz \right\| = \sqrt{\sum\limits_{k=1}\alpha_k^2\sigma_k^2},
\label{eqn:pert_growth}
\end{equation}
using the orthonormality of $\bfu_k$. If $\sigma_k > 1$, then the perturbation is said to be \textit{unstable} in the direction of the corresponding singular vector since the coefficient along this direction grows and small errors will be amplified. The perturbation is \textit{stable} in the directions of singular vectors that correspond to $\sigma_k < 1$.

Our approach is to study the stability properties of a Jacobian by inspecting its spectrum, i.e. the distribution of its singular values. We will take the number of singular values that are larger than 1 as a coarse measure of how unstable a system is, but as we will see in Section 3, the picture is more nuanced than that.

If the Jacobian is square, one can also use the eigenvalue decomposition to investigate stability properties. However, in this paper we restrict ourselves to the SVD.

\subsection{Related Work}
% Some prior works have studied the robustness of neural networks to perturbations in the input images within the scope of adversarial attacks. 

Analysis of the Jacobian of networks as a whole has been used in the study of their trainability. The distribution of singular values of the Jacobian during training is investigated in \cite{pennington2017resurrecting, pennington2018emergence,saxe2013exact} with the help of random matrix theory; they draw the conclusion that the networks can be trained more efficiently if the Jacobian is well-behaved, i.e. stable. In a similar spirit, \cite{tarnowski2018dynamical, xiao2018dynamical} apply random matrix theory and mean field theory to investigate the Jacobian in the limit of extreme deep and/or wide networks. These papers focus on the behavior of the Jacobian of (usually simplified) networks during training, whereas our objective is to understand the behavior of pre-trained networks as they are used in practice, and to this end we also primarily consider the Jacobians of the network components rather than the network as a whole.

% [NOT SURE IF THIS IS RELEVANT]
% Several measures of robustness are proposed and examined. \textit{Ensemble robustness} was presented in \cite{zahavy2016ensemble} and that paper claimed that it is important to the generalization of stochastic learning algorithms. Norms of the Jacobian (referred to \textit{input-output Jacobian} in these papers) are also used as a useful measure. \cite{sokolic2017robust} showed the connection between the robustness measured in norms of Jacobian and the generalization performance, and empirical support of this result is offered by \cite{zahavy2016ensemble}. That paper conducted experiments on a wide variety of neural networks trained on MNIST and CIFAR-10. They also tested a measure that is based on the change of linear region along trajectories between samples. 

\section{The Structure of ResNets and Their Linearizations}

ResNets are essentially the composition of a sequence of \textit{network stages}, each of which in turn is the composition of a sequence of \textit{residual units}. For standard ResNets, in a given stage we update the \textit{activation map} $\bfx_k$ using the update formula
\begin{equation}
    \bfx_{k+1} = \sigma\left(\bfx_k + \bfcalR_k\left(\bfx_k; \bftheta_k\right)\right),
    \label{eqn:residual_unit}
\end{equation}
where $\bfcalR_k\left(\bfx_k; \bftheta_k\right)$ is the $k$th residual block. The addition of $\bfx_k$ to the output of the residual block is referred to as a \textit{skip} or \textit{shortcut connection}, and $\sigma$ is a nonlinear activation function, usually the rectified linear unit (ReLU). All of these components combine to give the $k$th \textit{residual unit}. 

The network stages consist of a sequence residual units that take activation maps of the same size as their input, although these activation maps will differ in size between stages. Therefore, the first residual block in a stage has a slightly different structure than the subsequent "regular" residual blocks and down-samples the output from the previous stage into the appropriate size for the current stage. As we will see, the different architecture of this down-sampling unit leads to a substantially different spectrum of its Jacobian. For ResNet-56 and ResNet-110 trained on CIFAR-10, the two types of ResNets we use in this study, each stage has the same number of residual units (9 and 18, respectively).

The residual blocks $\bfcalR_k\left(\bfx_k; \bftheta_k\right)$ consist of convolution operators, batch normalizations and nonlinear activation functions. The weights $\bftheta_k$ are found by training the network and consist mostly of the weights required by the convolutional operators. We do not address the architecture of the residual blocks here, but there are different standard designs that are used in practice, depending on the depth of the network and the data set being studied. 

The linearization of the residual unit in~\eqref{eqn:residual_unit} is
\begin{equation*}
    \bfJ = \mathrm{diag}\left(\sigma'(\bfx_k + \bfcalR(\bfx_k))\right)\left(\bfI + \bfJ_{\bfcalR}\right),
\end{equation*}
where $\bfJ_{\bfcalR}$ is the linearization of the residual block. Note that the Jacobian depends directly on the activation map $\bfx_k$, which in turn depends on the input image. We implement the Jacobian as a neural network module that allows for the efficient computation of the product of the Jacobian with arbitrary vectors. These products are required when computing the SVD of a Jacobian iteratively; we in fact also require the \textit{transpose} of the Jacobian, but omit the details here.

Since the stages in a network are just compositions of residual units, the linearizations of the stages can be found by using the chain rule and the linearizations of the residual units.

\section{Experiments}

Our objective is to come closer to understanding the inner workings of pre-trained residual networks, or at least identify areas of future exploration that could prove to be fruitful in this regard, therefore most of our analysis will focus on the behavior of the residual units themselves, since these are the building blocks of the networks. 

\subsection{Setup}

Pre-trained ResNet-56 and ResNet-110 models trained on the CIFAR-10 data set were obtained from~\cite{semery2019}. Computation of the products of the Jacobians (and their transposes) with arbitrary vectors is implemented in PyTorch \cite{pytorch}. Eigenvalue and singular value decompositions were performed using NumPy and SciPy routines, which respectively use ARPACK and LAPACK to perform the computations. 

The following experiments were conducted on the CIFAR-10 test set \cite{cifar10} and run on Amazon AWS on \texttt{p3} and \texttt{g3} instance types. We have performed the experiments on a wide array of images, but show the results for only a few of those images here (the selected images are arbitrary and representative). We look only at results for ResNet-56 and ResNet-110 in this study.

\subsection{Spectra of the Jacobians of Residual Units and Network Stages }

\begin{figure}
\centering
\begin{tikzpicture}
  \node (img1) {\includegraphics[width=.29\linewidth]{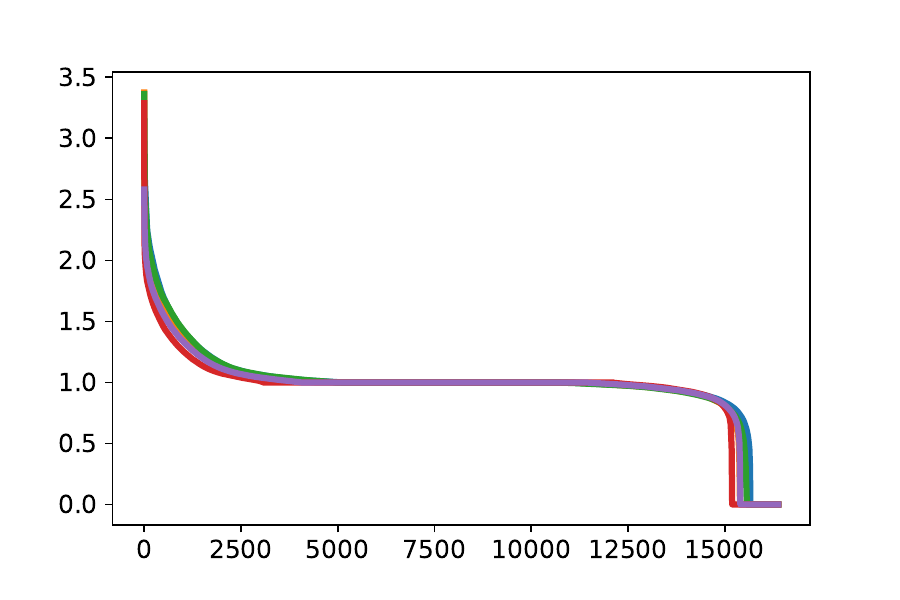}};
  \node (img2) at (img1.north east) [yshift=0.01cm] [xshift=-0.8cm]  {\includegraphics[width=.29\linewidth]{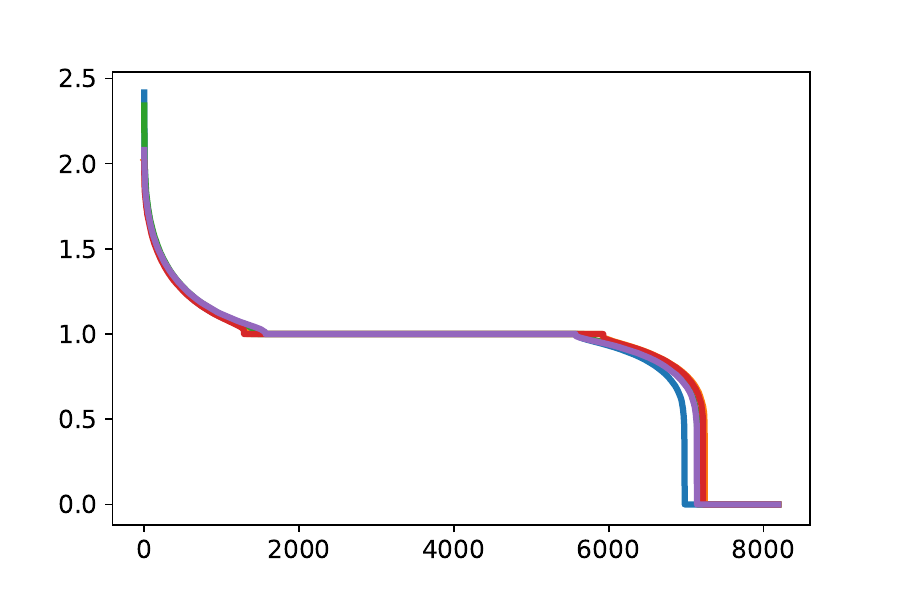}};
  \node (img3) at (img2.north east) [yshift=0.01cm] [xshift=-0.8cm] {\includegraphics[width=.29\linewidth]{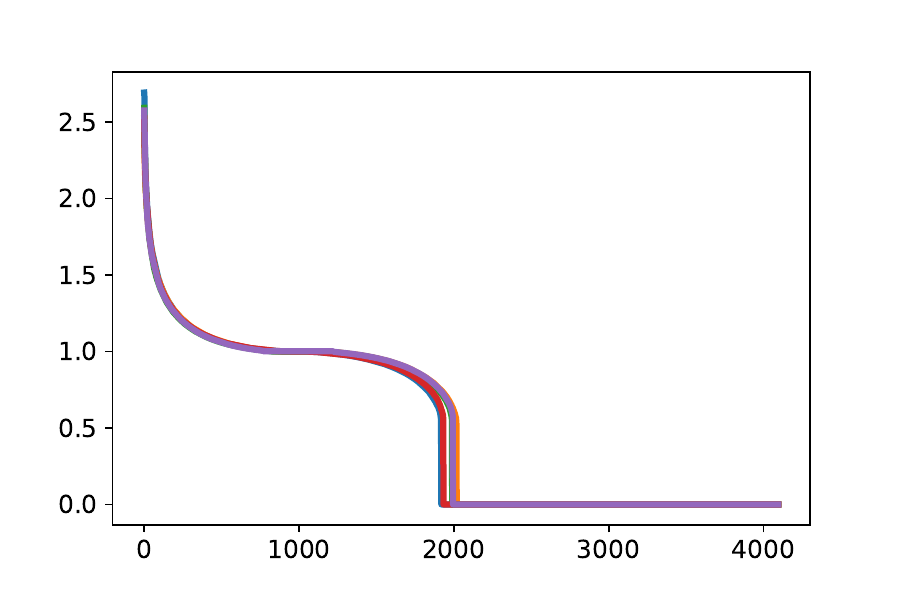}};
\end{tikzpicture}
% \hspace{-2cm}
% \begin{tikzpicture}
%   \node (img1) {\includegraphics[width=.33\linewidth]{images/cifar10_resnet56_00001_00005_stage1_unit004_clean_histogram}};
%   \node (img2) at (img1.north east) [yshift=0.2cm] [xshift=-0.6cm]  {\includegraphics[width=.33\linewidth]{images/cifar10_resnet56_00001_00005_stage2_unit004_clean_histogram}};
%   \node (img3) at (img2.north east) [yshift=0.2cm] [xshift=-0.6cm] {\includegraphics[width=.33\linewidth]{images/cifar10_resnet56_00001_00005_stage3_unit004_clean_histogram}};
% \end{tikzpicture} \\
\hspace{-0.5cm}
\begin{tikzpicture}
  \node (img1) {\includegraphics[width=.29\linewidth]{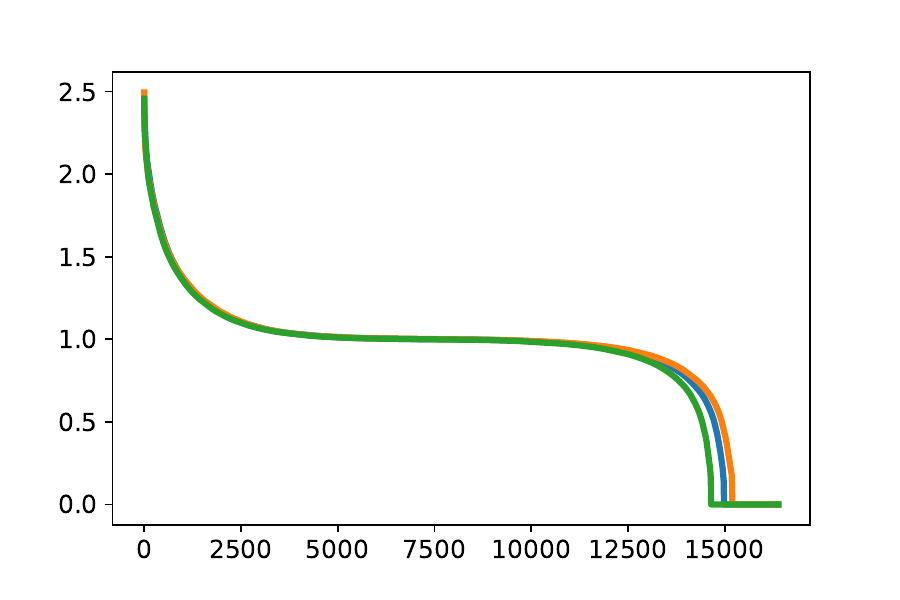}};
  \node (img2) at (img1.north east) [yshift=0.01cm] [xshift=-0.8cm]  {\includegraphics[width=.29\linewidth]{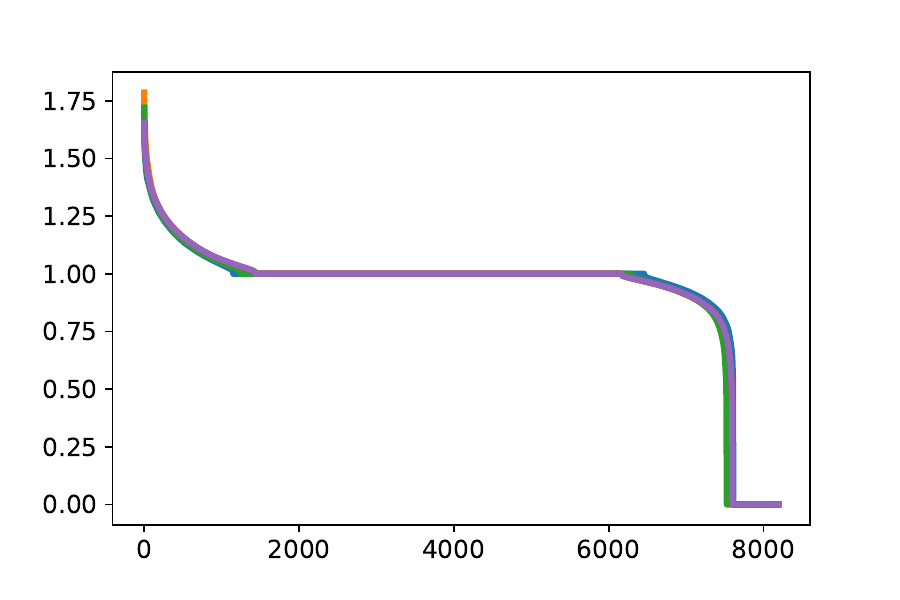}};
  \node (img3) at (img2.north east) [yshift=0.01cm] [xshift=-0.8cm] {\includegraphics[width=.29\linewidth]{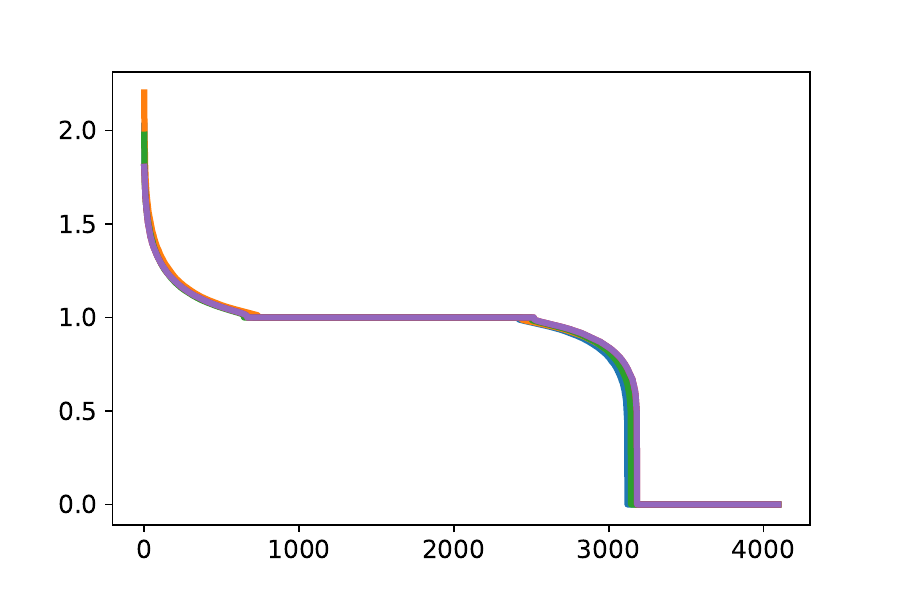}};
\end{tikzpicture} \\
% \hspace{-2cm}
% \begin{tikzpicture}
%   \node (img1) {\includegraphics[width=.33\linewidth]{images/cifar10_resnet110_00001_00005_stage1_unit004_clean_histogram}};
%   \node (img2) at (img1.north east) [yshift=0.2cm] [xshift=-0.6cm]  {\includegraphics[width=.33\linewidth]{images/cifar10_resnet110_00001_00005_stage2_unit004_clean_histogram}};
%   \node (img3) at (img2.north east) [yshift=0.2cm] [xshift=-0.6cm] {\includegraphics[width=.33\linewidth]{images/cifar10_resnet110_00001_00005_stage3_unit004_clean_histogram}};
% \end{tikzpicture}
\begin{tikzpicture}
  \node (img1) {\includegraphics[width=.29\linewidth]{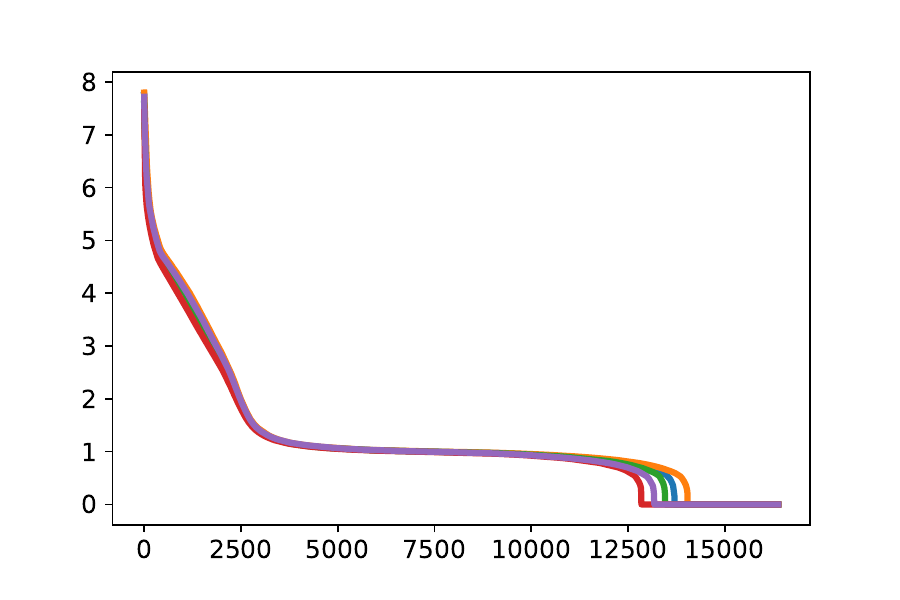}};
  \node (img2) at (img1.north east) [yshift=-0.25cm] [xshift=-0.8cm]  {\includegraphics[width=.29\linewidth]{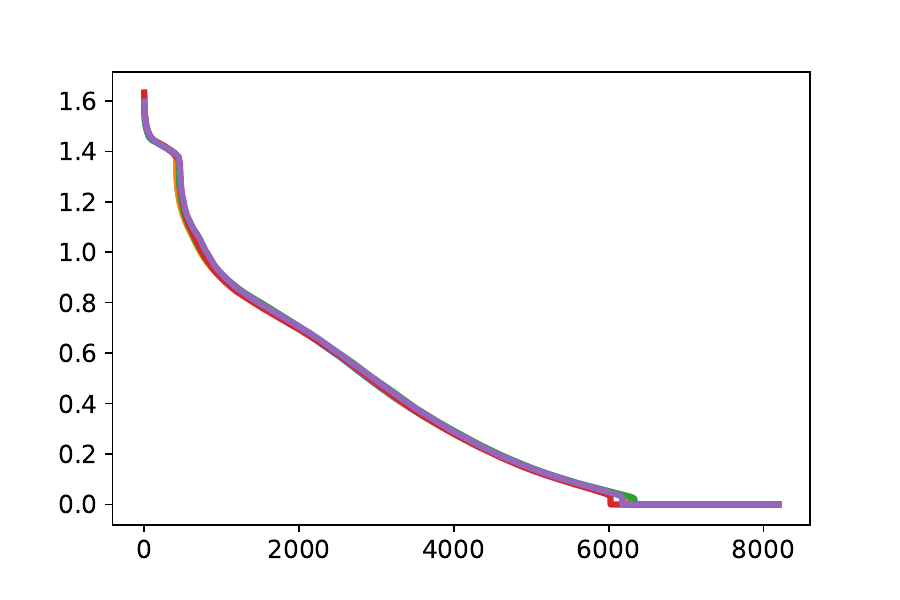}};
  \node (img3) at (img2.north east) [yshift=-0.25cm] [xshift=-0.8cm] {\includegraphics[width=.29\linewidth]{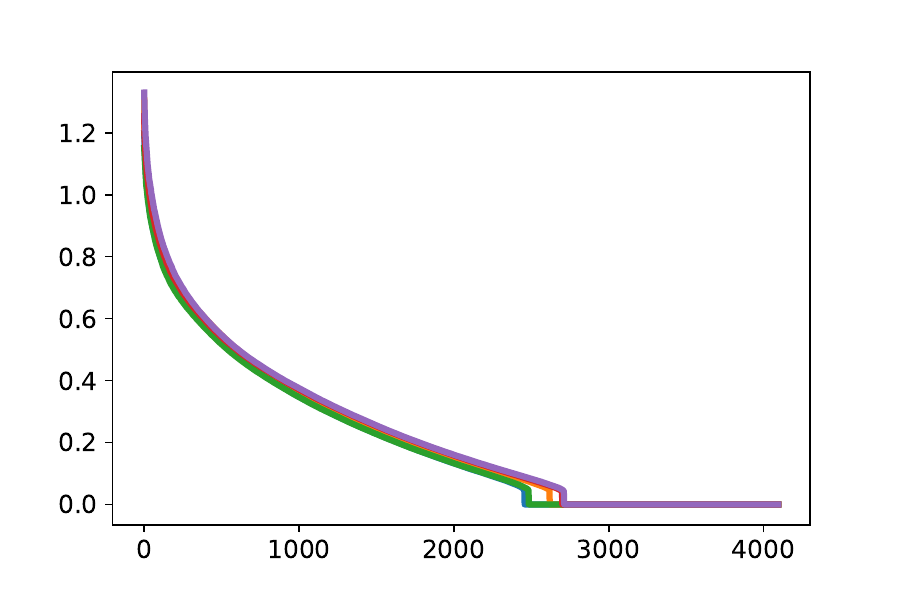}};
\end{tikzpicture}
\hspace{-0.5cm}
% \begin{tikzpicture}
%   \node (img1) {\includegraphics[width=.3\linewidth]{images/cifar10_resnet56_00001_00005_stage1_unit000_clean_histogram}};
%   \node (img2) at (img1.north east) [yshift=0.2cm] [xshift=-0.6cm]  {\includegraphics[width=.3\linewidth]{images/cifar10_resnet56_00001_00005_stage2_unit000_clean_histogram}};
%   \node (img3) at (img2.north east) [yshift=0.2cm] [xshift=-0.6cm] {\includegraphics[width=.3\linewidth]{images/cifar10_resnet56_00001_00005_stage3_unit000_clean_histogram}};
% \end{tikzpicture} \\
% \hspace{-2cm}
\begin{tikzpicture}
  \node (img1) {\includegraphics[width=.29\linewidth]{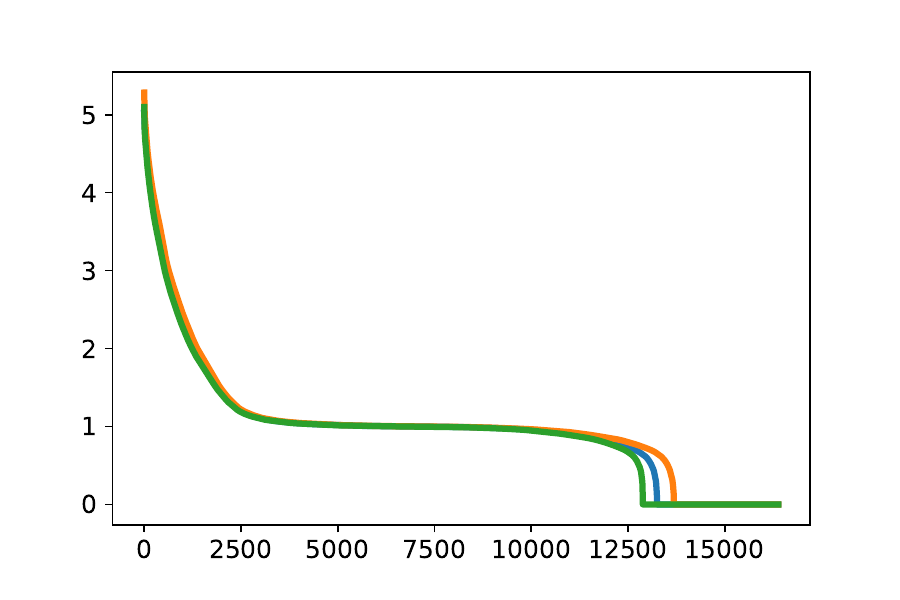}};
  \node (img2) at (img1.north east) [yshift=-0.25cm] [xshift=-0.8cm]  {\includegraphics[width=.29\linewidth]{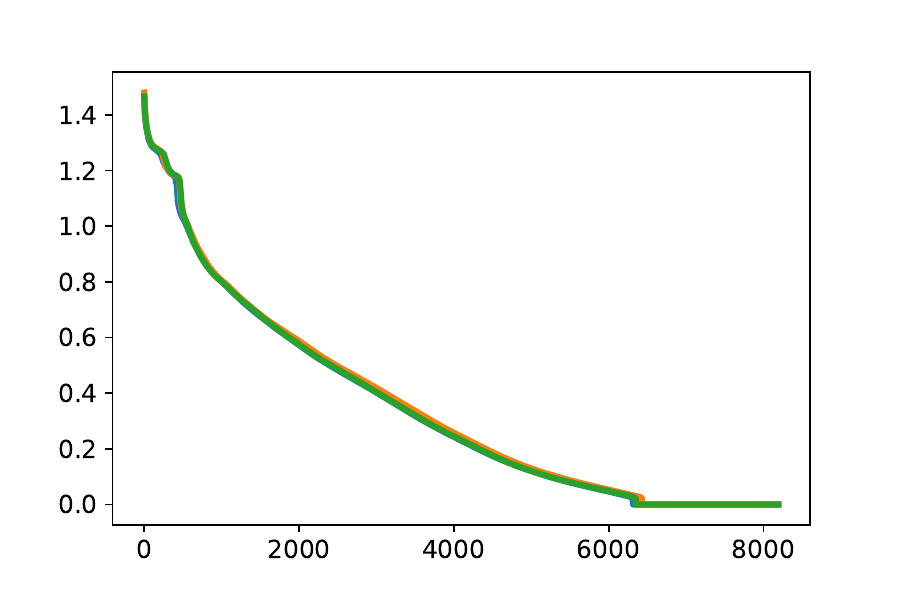}};
  \node (img3) at (img2.north east) [yshift=-0.25cm] [xshift=-0.8cm] {\includegraphics[width=.29\linewidth]{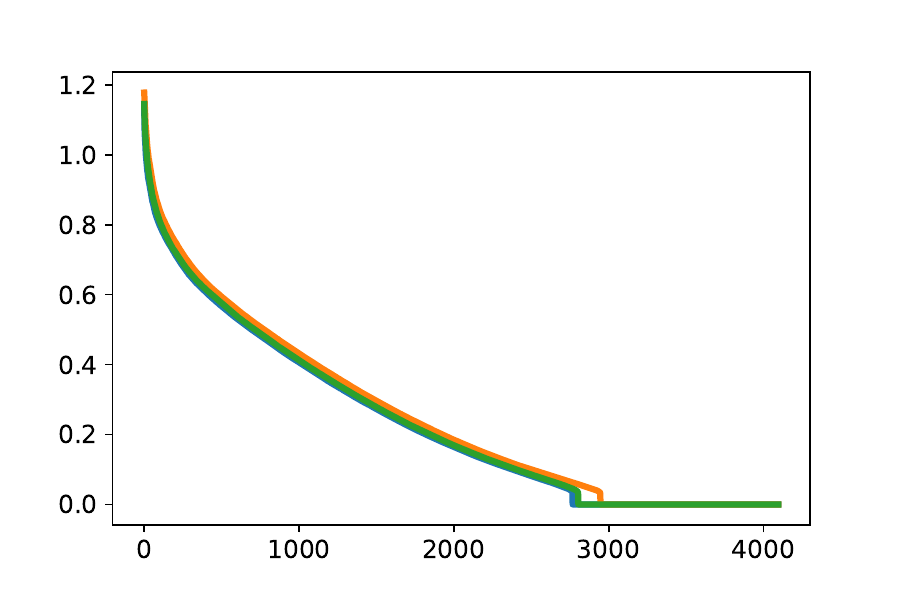}};
\end{tikzpicture} \\
% \hspace{-2cm}
% \begin{tikzpicture}
%   \node (img1) {\includegraphics[width=.3\linewidth]{images/cifar10_resnet110_00001_00005_stage1_unit004_clean_histogram}};
%   \node (img2) at (img1.north east) [yshift=0.2cm] [xshift=-0.6cm]  {\includegraphics[width=.3\linewidth]{images/cifar10_resnet110_00001_00005_stage2_unit004_clean_histogram}};
%   \node (img3) at (img2.north east) [yshift=0.2cm] [xshift=-0.6cm] {\includegraphics[width=.3\linewidth]{images/cifar10_resnet110_00001_00005_stage3_unit004_clean_histogram}};
% \end{tikzpicture}
\begin{tikzpicture}
  \node (img1) {\includegraphics[width=.29\linewidth]{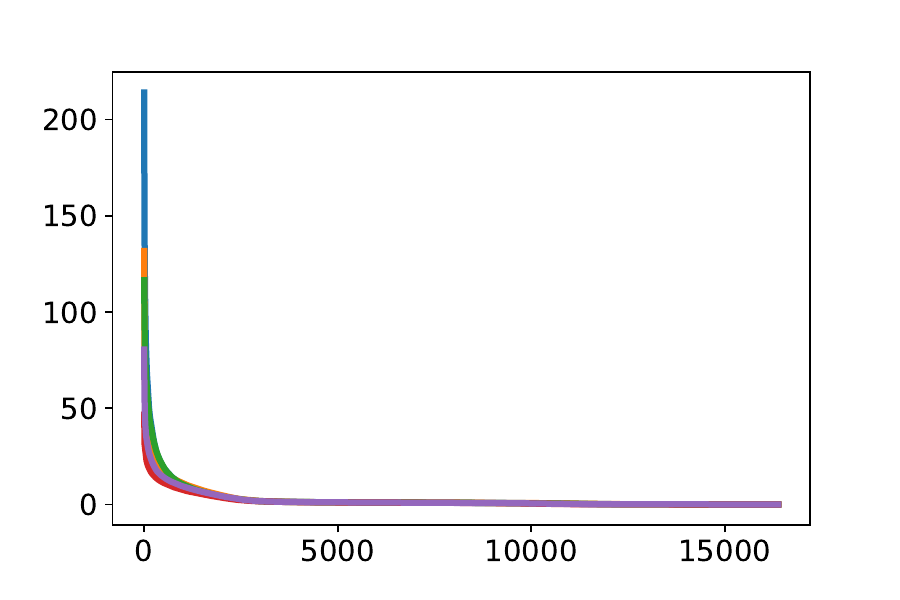}};
  \node (img2) at (img1.north east) [yshift=-0.28cm] [xshift=-0.8cm]  {\includegraphics[width=.29\linewidth]{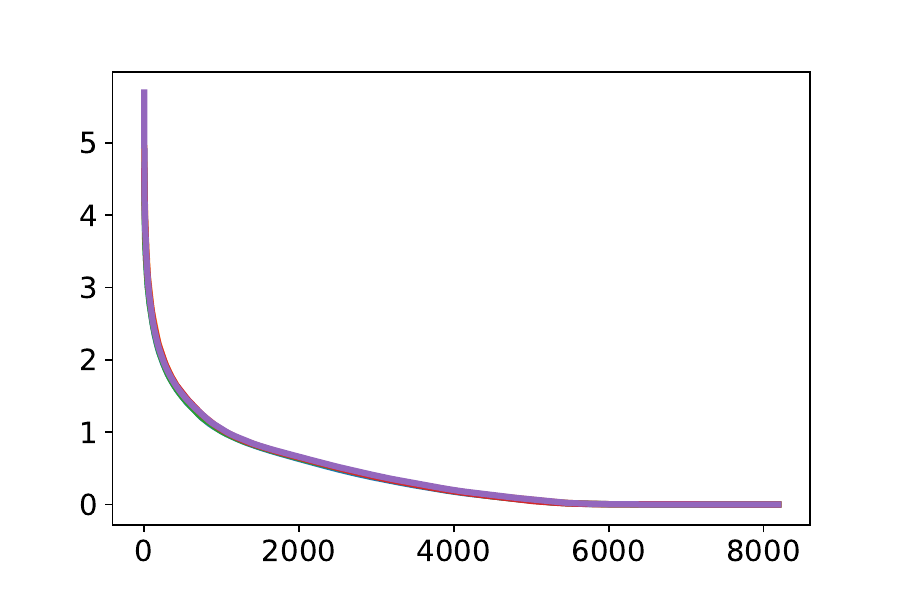}};
  \node (img3) at (img2.north east) [yshift=-0.28cm] [xshift=-0.8cm] {\includegraphics[width=.29\linewidth]{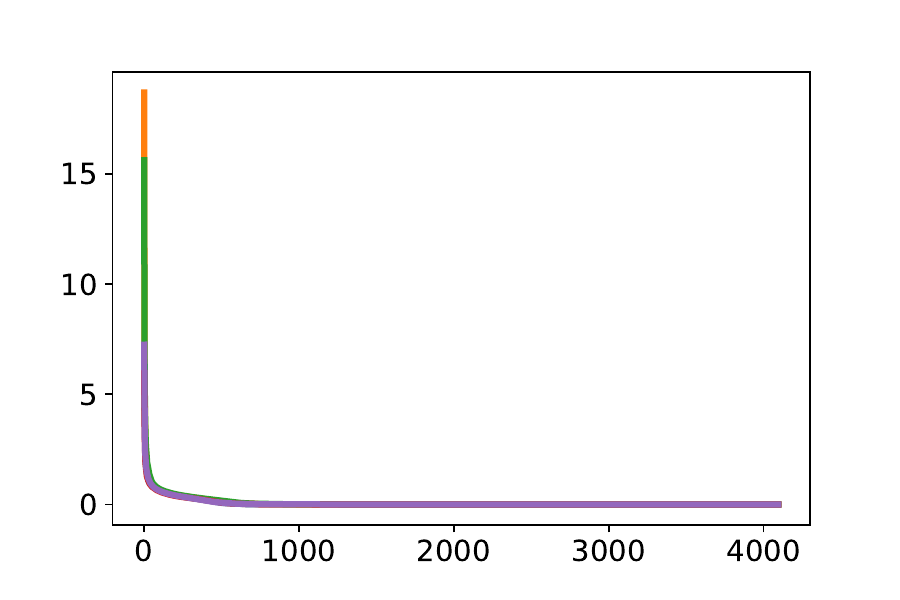}};
\end{tikzpicture}
\hspace{-0.5cm}
\begin{tikzpicture}
  \node (img1) {\includegraphics[width=.29\linewidth]{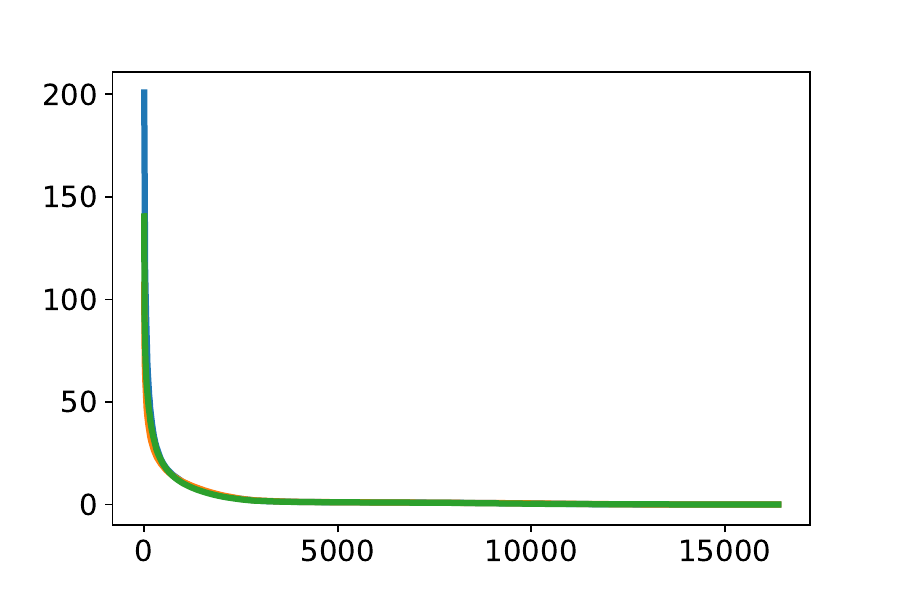}};
  \node (img2) at (img1.north east) [yshift=-0.28cm] [xshift=-0.8cm]  {\includegraphics[width=.29\linewidth]{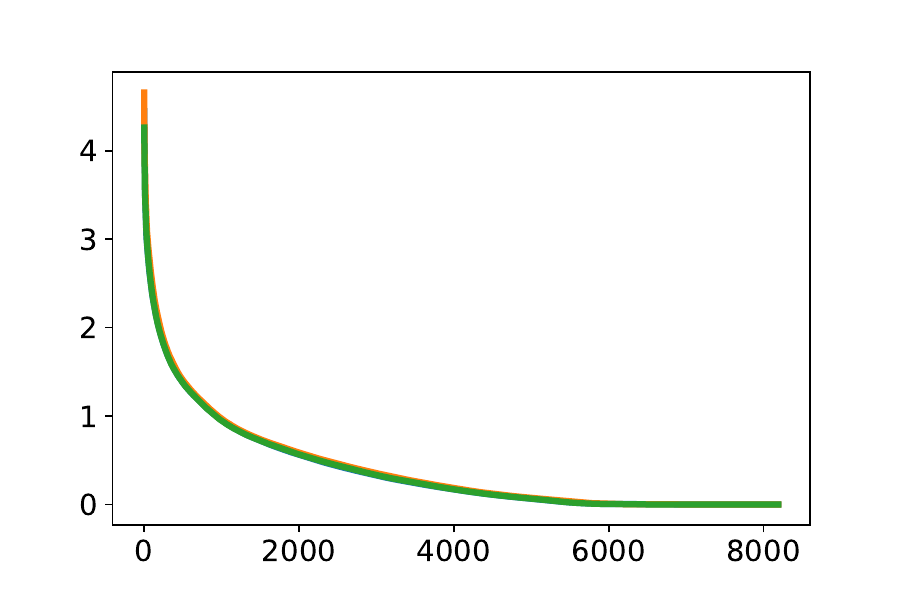}};
  \node (img3) at (img2.north east) [yshift=-0.28cm] [xshift=-0.8cm] {\includegraphics[width=.29\linewidth]{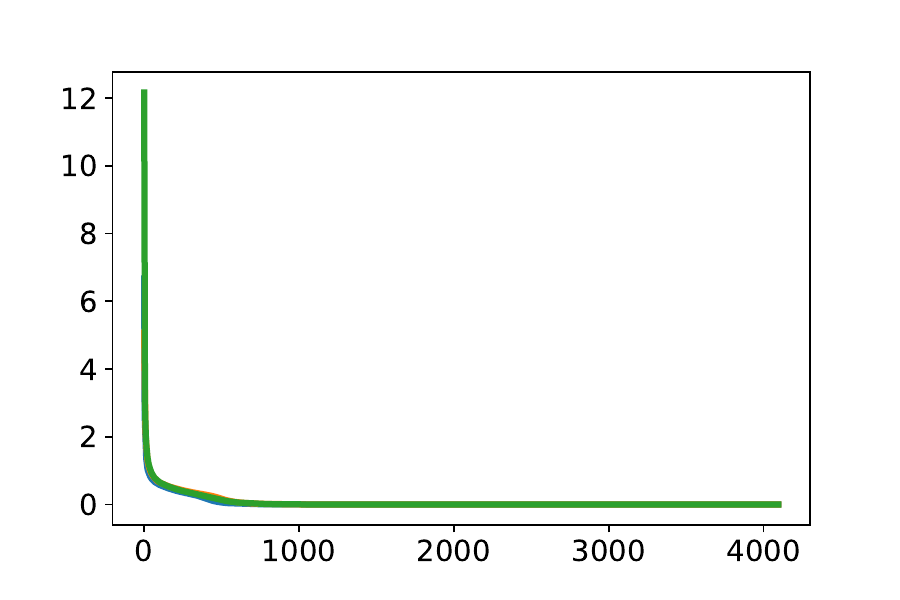}};
\end{tikzpicture}
\caption{Scree plots of the singular values of the Jacobians of different network components. Within each set of three overlapping images, the first stage is bottom left and the third stage is top right. The spectra for ResNet-56 are on the left, and the results for ResNet-110 on the right. To illustrate what little effect different images have on the spectra, singular values are shown for five different images from the CIFAR-10 test set overlayed on top of each other. \\
\textit{Top row}: Spectra corresponding to the fifth residual unit in each of the network stages. Notice that the Jacobians have high rank and that a large proportion of singular values are 1, represented by the plateau that is evident in the plots. The plots are representative for other residual units in the network, except for the down-sampling residual units. \\
\textit{Middle row}: Spectra corresponding to the initial, down-sampling residual units in the network stages. The first stage still has singular values with a mean of 1, but the plateau is less pronounced. The second and third stages do not exhibit the same singular value distributions as other residual units in the network. \\
\textit{Bottom row}: Spectra corresponding to the network stages. Notice the absence of the plateaus and a small number of very large singular values. 
% Singular values of the down-sampling residual unit in each of the network stages of ResNet-56 (top row) and Resnet-110 (bottom row). For ResNet-56, singular values are for 5 different images from the CIFAR-10 test set overlayed on top of each other; we used three different images for the Resnet-110 plots. The scree plot is shown on the left and a histogram of the singular values is shown on the right. Apart from the first stage, the spectra do not have the plateau that regular residual units have. In the first stage the plateau is less pronounced than for regular residual units.
% Singular values of the network stages of ResNet-56 (left) and Resnet-110 (right). For ResNet-56, singular values are for 5 different images from the CIFAR-10 test set overlayed on top of each other; we used three different images for the Resnet-110 plots. The spectra of the two ResNet architectures are very similar.
}
\label{fig:spectra}
\end{figure}

Representative spectra of the Jacobians of residual units are shown in Figure~\ref{fig:spectra}. In this case we show the fifth residual unit of each of the stages of ResNet-56 and ResNet-110. To illustrate their similarity across different images, we have overlayed the spectra of five images on top of each other. In every case there is a large number of singular values at 1, which shows up as a plateau in the scree plots. There is a non-negligible number of singular values above 1 and therefore the residual units are unstable in the directions of the corresponding singular vectors. Notice the lack of low-rank structure exhibited by the singular values, except for the third stage of ResNet-56. We explore this more closely below. We mention that it was proposed in~\cite{glorot2010} that learning in deep feed-forward neural networks can be improved by keeping the mean singular value of the Jacobian of each residual unit close to 1, so it is tempting to argue that this singular value distribution is the result of an efficient training procedure. However, as we explore briefly in Section 3.4, having a singular value distribution with a mean of 1 appears to be due to the structure of the residual units themselves.

% [NOT NEEDED?] Since the Jacobians of the standard residual units are square, we can also perform an eigenvalue decomposition on them. Recall that for stability to hold, we require that their magnitude is less than 1, but we observe a clustering around $1$, see Figure~\ref{fig:standard_unit_scatter_plots}. In our experiments we generally see interesting structures of the eigenvalue distributions in the first stage, but these structures have little effect on the actual stability properties.

% \begin{figure}
% \includegraphics[width=.32\linewidth]{images/cifar10_resnet56_00001_00005_stage1_unit004_clean_scatter}
% \includegraphics[width=.32\linewidth]{images/cifar10_resnet56_00001_00005_stage2_unit004_clean_scatter}
% \includegraphics[width=.32\linewidth]{images/cifar10_resnet56_00001_00005_stage3_unit004_clean_scatter}
% \caption{Scatter plots of the eigenvalue distributions in the complex plane of the fourth unit in each of the three stages (in sequence) in ResNet-56. We have overlaid the eigenvalue plots of five different images. The lavender disk indicates the stability region of the Jacobian, i.e. perturbations grow in the directions of the eigenvectors corresponding to eigenvalues that lie outside of the disk. }
% \label{fig:standard_unit_scatter_plots}
% \end{figure}

\begin{figure}
\centering
\includegraphics[width=.9\linewidth]{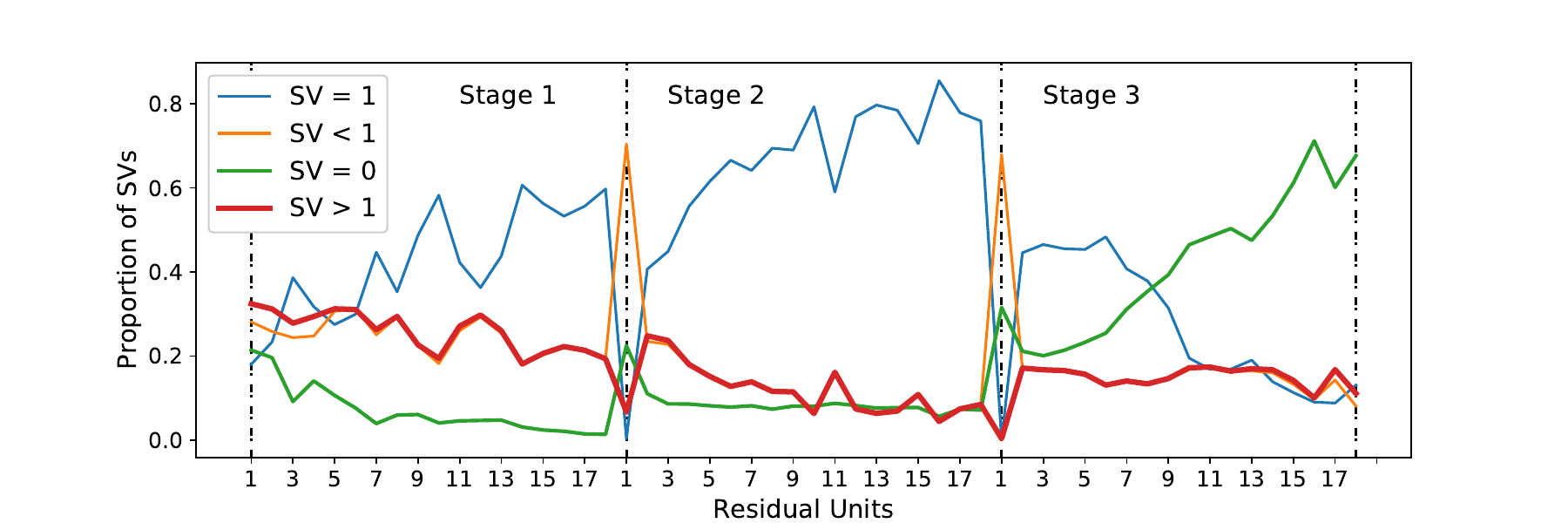}
\caption{Distribution of singular values that are larger than 1.01 (bold red), between 0.99 and 1.01 (blue), between zero and 0.99 (orange), and zero (green) for each of the three stages in ResNet-110 for an arbitrary input image.}
\label{fig:bins}
\end{figure}

It is interesting to observe how the distribution of singular values changes as we move through the network. In Figure~\ref{fig:bins} we show the proportion of singular values that are larger than 1, between 0.99 and 1.01, between 0.0 and 0.99, and at 0.0. We note the following:
\begin{itemize}
    \item The number of singular values that are greater than 1 decreases slightly as we move through the network, indicating that the residual blocks become increasingly stable by our measure, although not dramatically.
    \item The number of singular values at 1 increases throughout the first two stages and decreases drastically in the third stage.
    \item The number of singular values at 0 decreases slightly during the first two stages, which means that the ranks of the Jacobians start off large and become even bigger. On the other hand, there is a steep decrease in the number of non-zero singular values in the third stage, so that the ranks of the Jacobians in this stage decrease as we propagate through it.
\end{itemize}
The evolution of the spectra of the Jacobians of the residual units in the third stage behaves substantially different from that of the spectra in the first two stages. In all three stages we see the plateau of height 1 in the scree plots, but in the third stage it loses its width as we move through the stage. 

As mentioned in Section 2, the first residual unit in a stage resizes the activation maps to a size required by that stage. We therefore expect the residual units to have slightly different stability properties and this is indeed the case, see Figure~\ref{fig:spectra}. In this case the first stage still has a large number of singular values around 1, but the plateau is less pronounced. Stages 2 and 3 show markedly different behavior in that they do not have a significant portion of singular values close to 1. The spectra of the Jacobians of the network stages are shown in Figure~\ref{fig:spectra} and in this case the singular values have very different distributions than the singular values of the residual units that make up the network stages. Recall that the Jacobians of the stages are obtained using the chain rule and depend directly on the Jacobians of the residual units, so it will be interesting to further explore how the spectra corresponding to the stages depend on those of the residual units, although this lies outside of the scope of the present study. Again, there is no significant difference between the spectra corresponding to different input images, or between ResNet-56 and ResNet-110.

\subsection{Propagated Perturbations}

\begin{figure}
\centering
\includegraphics[width=.9\linewidth]{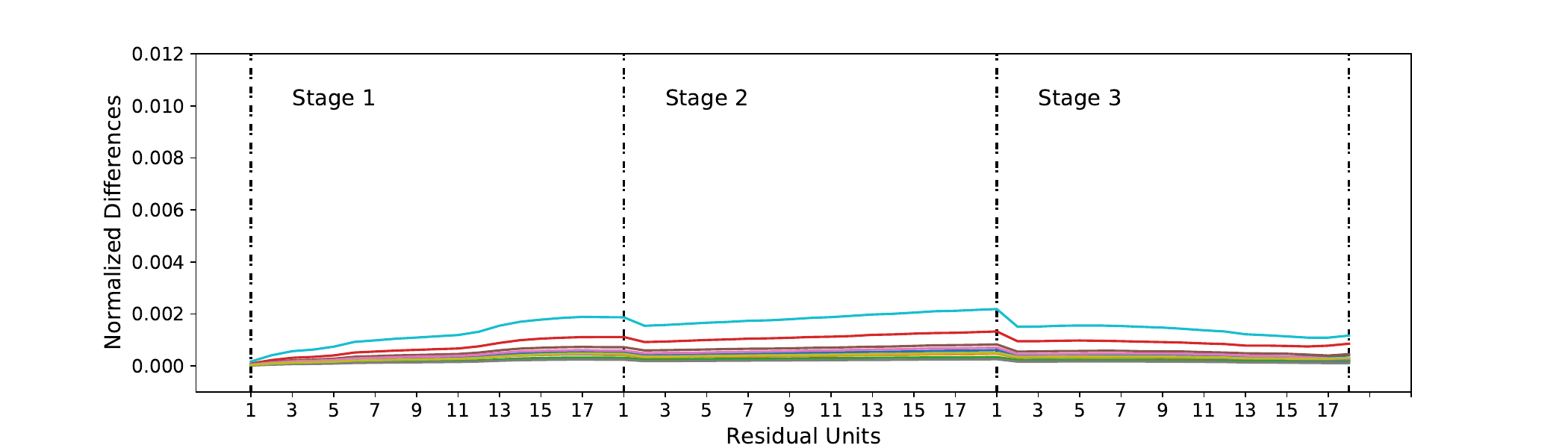} \\[-1.3em]
\includegraphics[width=.9\linewidth, trim={0 0 0 3.2em},clip]{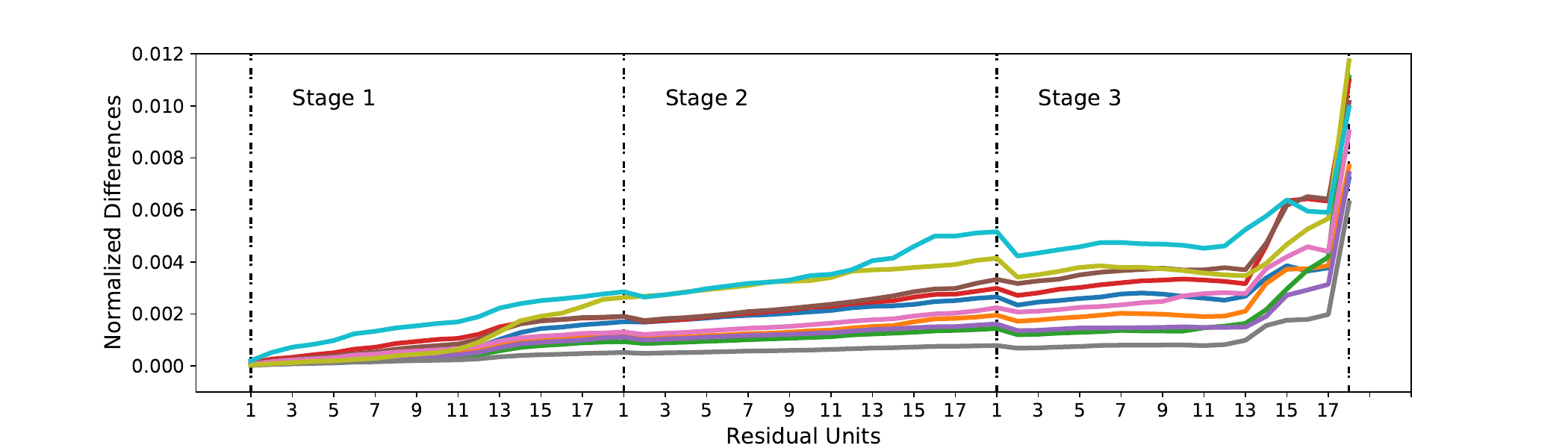}
\caption{Normalized magnitudes of differences between activation maps of clean images and randomly perturbed (top) and adversarially perturbed (bottom) images for each residual unit in ResNet-110, for ten different images in the CIFAR-10 test set.}
\label{fig:diffs}
\end{figure}

The magnitudes of the differences between the activation maps corresponding to clean images and those of perturbations of those images are shown in Figure~\ref{fig:diffs}, for ten different images of the CIFAR-10 test set. We will refer to these differences as propagated perturbations, since they correspond to the differences between the activation maps of perturbed images and those of the clean ones. The magnitudes shown in~\ref{fig:diffs} have been normalized with respect to the sizes of the activation maps, since these change between the stages.

We consider two types of perturbations here: random perturbations, where we add noise generated from a uniform distribution to an image, and adversarial perturbations, where we generate a perturbation using the FGSM attack method~\cite{goodfellow2014explaining, kurakin2016adversarial} that causes the previously correctly-identified image to be misclassified. We scale the random perturbations to have the same magnitude as the adversarial perturbations. The magnitudes of the propagated perturbations for the randomly perturbed image are on top, and those for the adversarially perturbed image are on the bottom.

In the first two stages, as we would expect from the fact that the residual units are unstable, the perturbations mostly increase. The propagated perturbations of adversarial images increase faster than those of randomly perturbed images. There is significantly different behavior in the third stage, where the propagated perturbations actually decrease for randomly perturbed images, but there is a sharp increase in the propagated perturbation towards the end of the third stage for adversarial images.

\begin{figure}
\centering
\includegraphics[width=.9\linewidth]{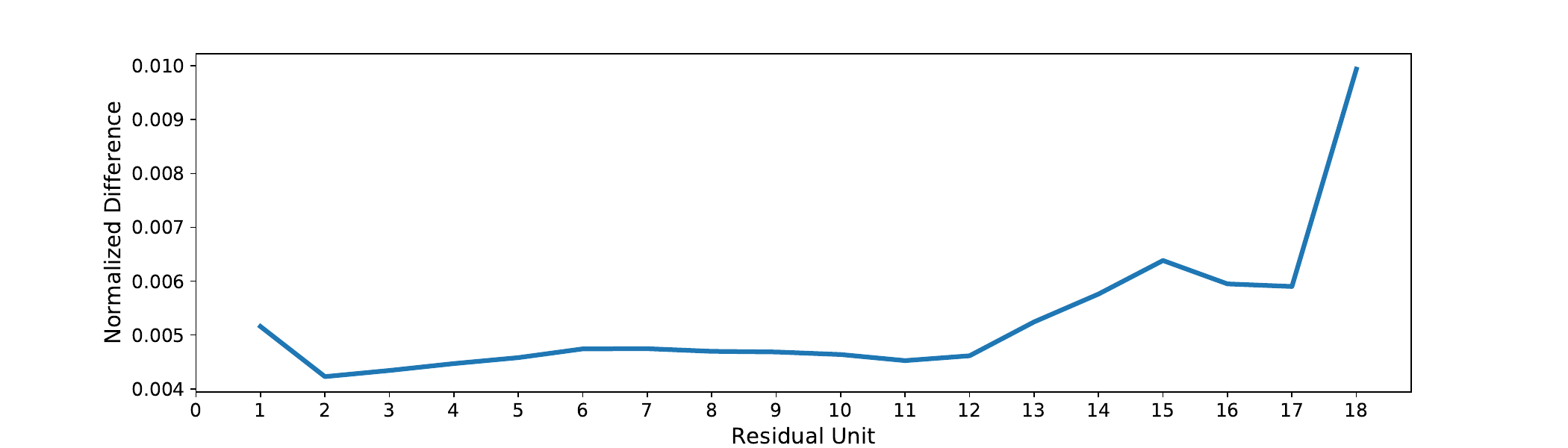} \\[-1.3em]
\includegraphics[width=.9\linewidth, trim={0 0 0 3.2em},clip]{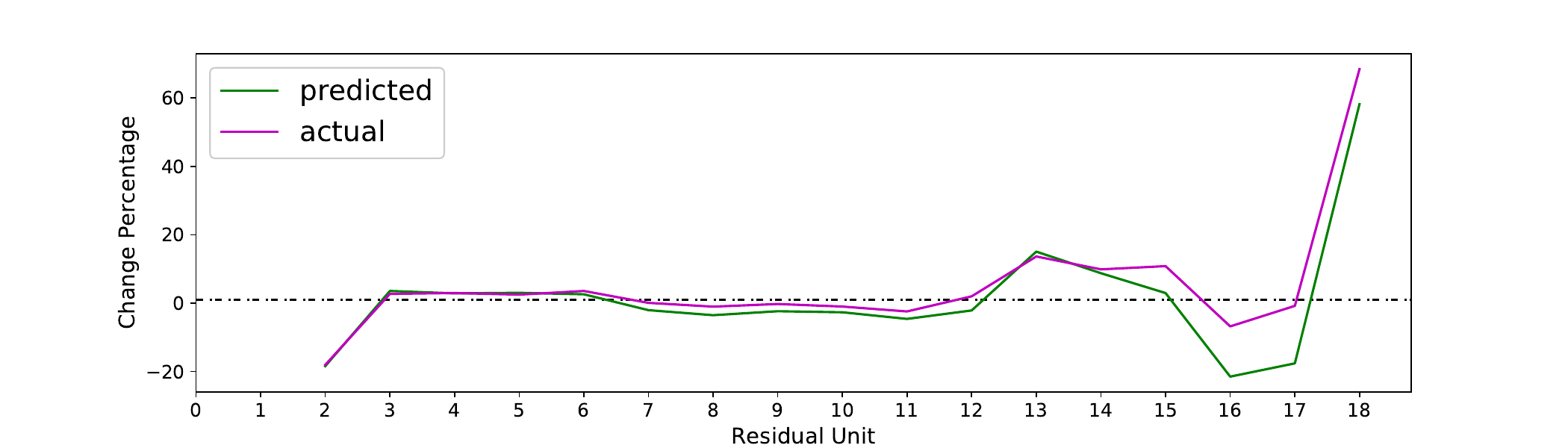} \\
\includegraphics[width=.257\linewidth, trim={3em 0 4.33em 2em},clip]{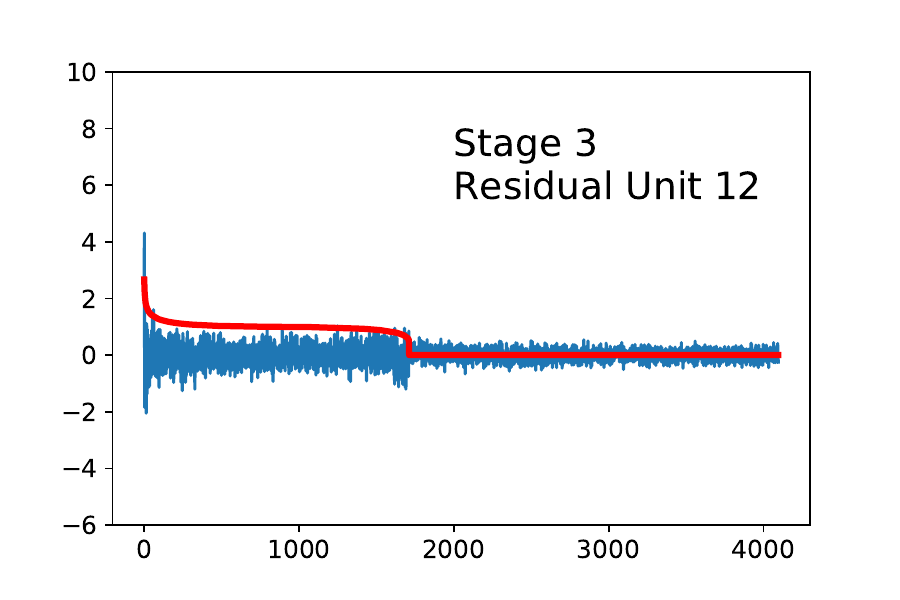}
\includegraphics[width=.241\linewidth, trim={5.37em 0 4.33em 2em},clip]{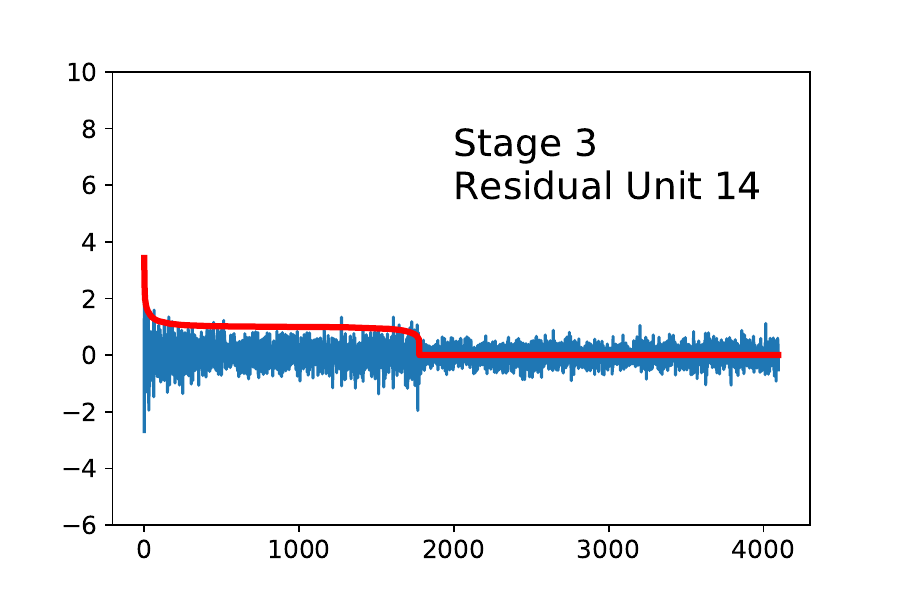}
\includegraphics[width=.241\linewidth, trim={5.37em 0 4.33em 2em},clip]{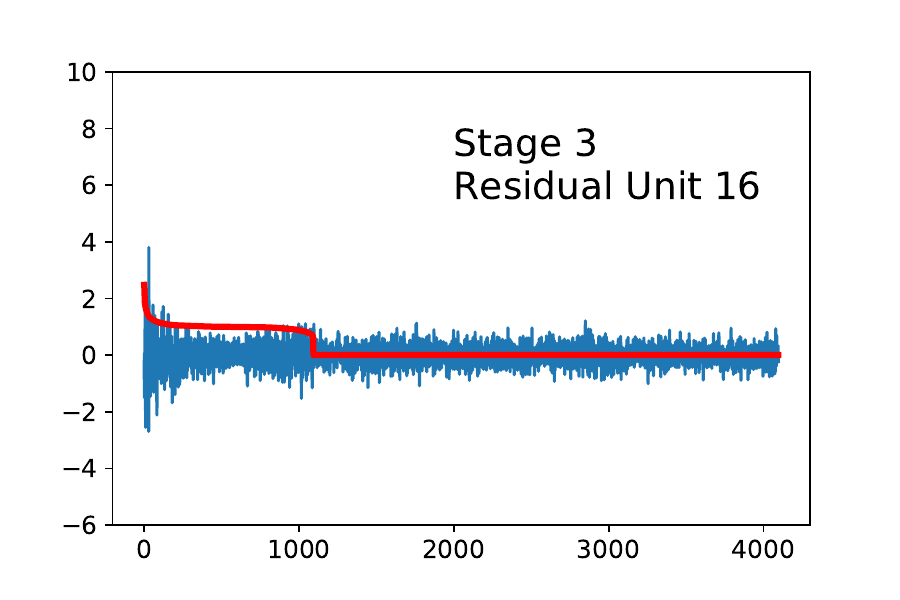}
\includegraphics[width=.241\linewidth, trim={5.37em 0 4.33em 2em},clip]{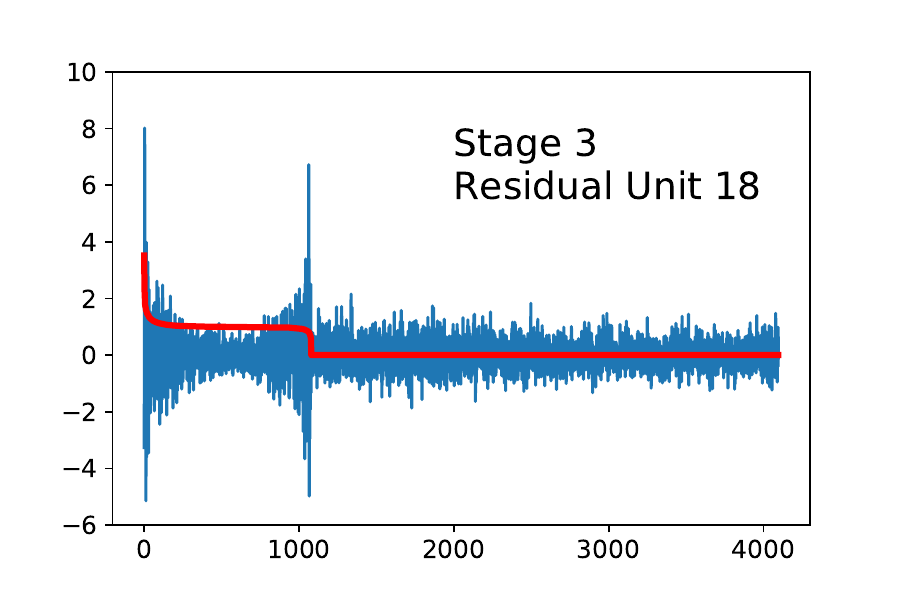}
\caption{\textit{Top}: Differences between activation maps of clean images and randomly perturbed images for each residual unit in the third stage of ResNet-110. Differences are normalized with respect to the size of the activation maps. \\
\textit{Middle}: Predicted and actual changes of the differences between the clean and perturbed activation maps for each residual unit, relative to the difference between the maps in the prior residual unit. See the text for a discussion. \\
\textit{Bottom}: Projections of the propagated perturbations onto the right singular vectors of the Jacobian of residual units 12, 14, 16, and 18 (left to right) in the third stage, with the corresponding singular values superimposed in red. Notice that the projections corresponding to the largest singular values are substantially larger in the later residual units, indicating that the perturbations are more closer oriented with these singular vectors than was the case in earlier residual units. This provides an explanation why there is a sudden increase in the perturbation at the end of the stage. }
\label{fig:stage3_diffs}
\end{figure}

Let us examine the adversarial perturbations in third stage more closely; see Figure~\ref{fig:stage3_diffs}. It may seem contradictory at first that the difference should grow so dramatically towards the end of the stage, given that Figure~\ref{fig:bins} shows that there are fewer singular values greater than 1 at this point of the network, which means that there are fewer directions along which the propagated perturbations can grow. This implies that the perturbations must be lining up along the directions defined by the singular vectors corresponding to the largest singular values.

Recall~\eqref{eqn:pert_growth}. Since $\alpha_i = \langle \delta \bfx_k, \bfv_i \rangle$, it is easy to compute the predicted change of the magnitudes of the perturbations. The larger the value of $\alpha_i$, the more closely aligned $\delta \bfx_k$ is with the singular vector $\bfv_i$. A simple calculation then allows us to compute predicted change in the magnitude of the perturbations and compare it with the actual change, as shown in the middle row of Figure~\ref{fig:stage3_diffs}. The final row of Figure~\ref{fig:stage3_diffs} shows the values of $\alpha_i$ for different residual units, with the singular values overlayed. Notice that the perturbation is much closer aligned with the singular vector corresponding to the largest singular value of the final residual unit than is the case elsewhere in the network. Not shown here, but this is not the case for random perturbations.

\subsection{The Effect of Unit Architecture and Weights on the Spectra of Residual Units}

We have seen above that there is little difference in the spectra of the residual units due to the images that are being fed into the network. It is therefore natural to ask what some of the factors are that influence the spectra, and while a comprehensive study of this lies outside the scope of this paper and will be addressed in future work, we will briefly outline some factors we have found to have an effect.

\subsubsection{Scaling the Skip Connections}

The architecture of a residual unit is expected to have a significant impact on the singular values of the unit's Jacobian. A simple illustration of this is to scale the skip connection with some arbitrary parameter $\alpha$, giving the update equation
\begin{equation*}
    \bfx_{k+1} = \sigma\left(\alpha\bfx_k + \bfcalR_k\left(\bfx_k\right)\right).
    \label{eqn:resnet_update_formula_scaled}
\end{equation*}
\begin{figure}
\centering
\includegraphics[width=.36\linewidth]{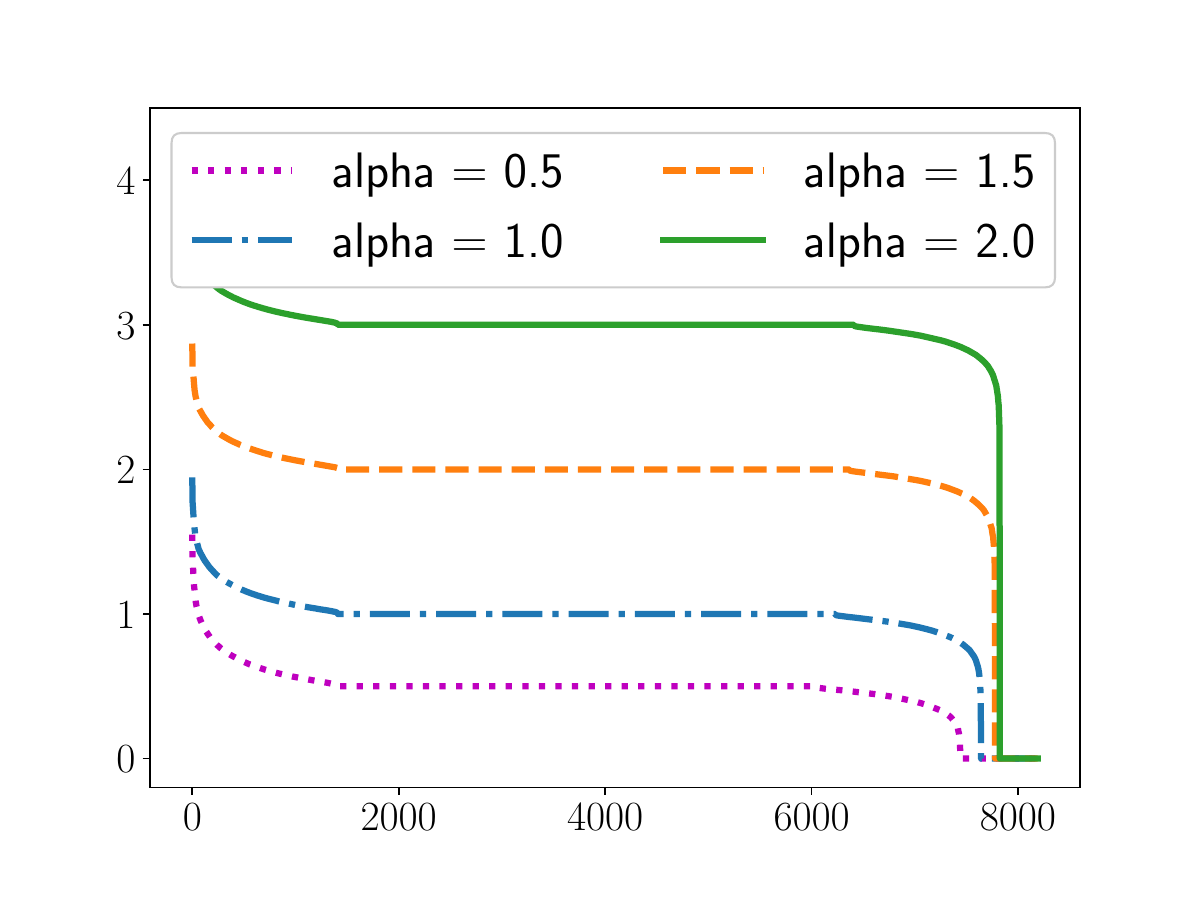}
\includegraphics[width=.63\linewidth]{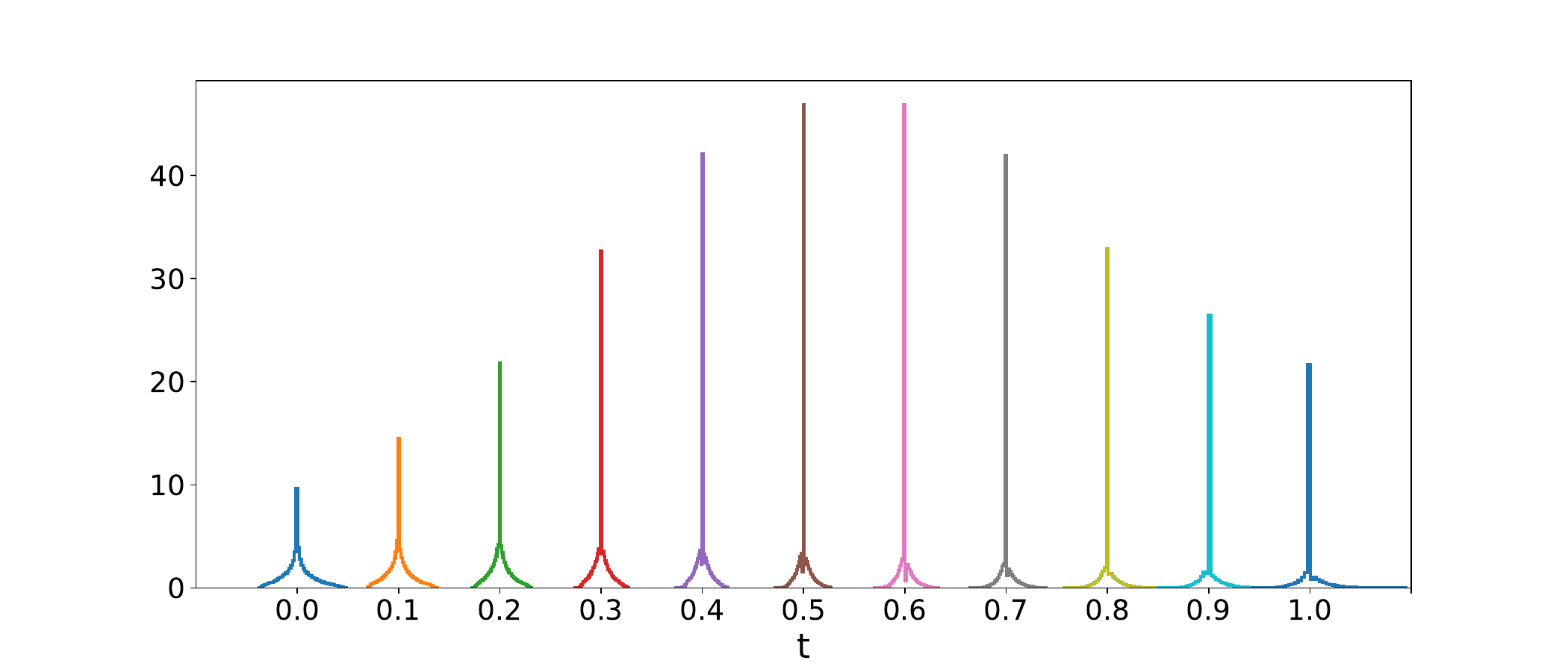}
\caption{\textit{Left}: Spectra for different scalings of the skip connection in a residual unit. The plateau attains the value of the scaling value $\alpha$, but the entire spectrum is not being scaled by $\alpha$. \\
\textit{Right}: Histograms of the singular value distributions for different values of $t$ in~\eqref{eqn:convex_weights}.
%  For $t=0$, the weights are pure noise and the singular values still have a mean of $1$, although relatively few of the singular values actually are 1. As $t$ increases more singular values attain the value of 1, but, interestingly, this number starts to decrease again once we move closer to the trained weights $\bftheta$.
 }
\label{fig:scaled_spectrum}
\end{figure}
The effect of this is seen on the left in Figure~\ref{fig:scaled_spectrum}, where we have computed the spectra for an arbitrary residual unit that has a scaled skip connection. Scaling has the effect of moving the plateau of the spectrum to the scaling parameter $\alpha$. What effect this has on the spectra of the network stages, or the network as a whole, will be investigated at a later date.

\subsection{Weights of Residual Unit Convolution Operators}

For regular residual units (i.e. residual units that do not down-sample the activation maps at the beginning of the stage), we have discussed the plateau that appears to always be present and that it changes in width in the third network stage.

We are interested in seeing how the distribution of the singular values changes as we change the weights of the convolutional operators in the residual units. For this, we replace existing pre-trained weights $\bftheta$ in an arbitrary residual unit with a convex combination of $\bftheta$ and pure Gaussian noise $\bfeta$ that has been scaled to have the same norm as $\bftheta$:
\begin{equation}
    \hat{\bftheta}(t) = t \bftheta + (1-t) \bfeta, \qquad 0 \leq t \leq 1.
    \label{eqn:convex_weights}
\end{equation}
The results are shown in the histograms on the right in Figure~\ref{fig:scaled_spectrum}. For pure noise $\bfeta$ ($t=0$), the plateau at $1$ is absent and the singular values are more spread out. The plateau slowly takes shape as $t$ increases, until it reaches a maximum for $t=0.5$, after which is loses some of its width and the singular values begin to spread out again. It will be interesting to explore this topic more in future work.

\section{Discussion}

Treating pre-trained ResNets as nonlinear systems allows us to use linearization to perform an empirical stability analysis, which we have done for ResNet-56 and ResNet-110 trained on the CIFAR-10 data set. We have found that most singular values of the Jacobians of residual units lie close to 1, which shows up as a plateau in the scree plots.

An in-depth analysis of our observations lies beyond the scope of this paper, but there are many observations that need to be investigated further. We have seen that the differences in the activation maps for clean and corresponding adversarial images differ significantly from those of clean and randomly perturbed images. This is particularly the case at the end of the final stage in the network, and while we showed that linearization provides a good estimate of how much the magnitude of the perturbation will change in the subsequent residual unit, a deeper understanding of why this is happening is still missing. In particular, how much of a role the attack \textit{method} plays is still unclear; it appears that the adversarial perturbations generated by FGSM have the effect of perturbing the final activation maps most significantly, but it remains to be seen if this is the case in general.

We have also seen that there is a significant dependence of the spectra on the architecture of residual units, which is in stark contrast to the lack of dependence on the input images. This will also be the subject of study in future work.

% \clearpage

% \section*{References}
{\small
\bibliographystyle{ieee}
\bibliography{resnet_linearizations_neurips2019}
}

\newpage

\appendix

\section{Linearizations of \texttt{BasicBlock} Residual Units}

\begin{figure}[H]
\centering
\includegraphics[width=.83\linewidth]{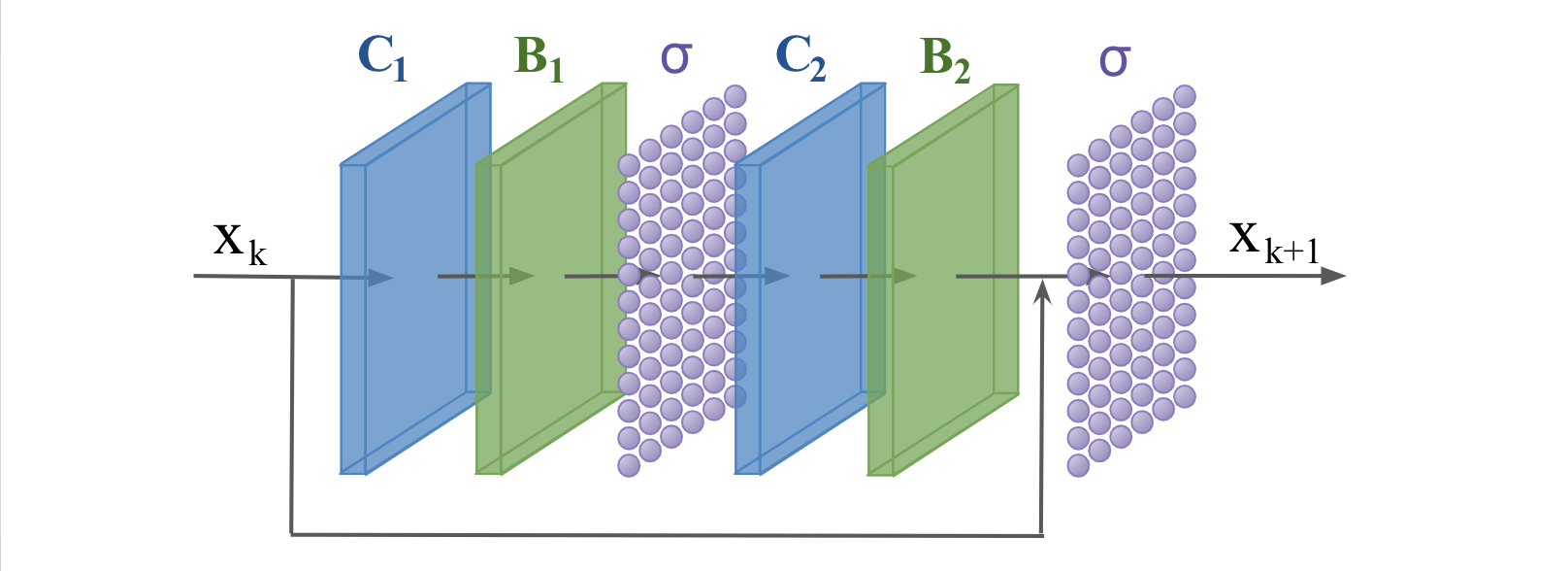}
\caption{Schematic representation of \texttt{BasicBlock} residual unit architecture.}
\label{fig:basicblock}
\end{figure}

The two ResNet architectures we use in this study, ResNet-56 and ResNet-110, both use the \texttt{BasicBlock} architecture for their residual blocks, shown in Figure~\ref{fig:basicblock}. Here he provide a couple of additional details on the internal structure of \texttt{BasicBlock} and its linearization.

The general update formula for ResNets was given in~\eqref{eqn:residual_unit}
\begin{equation*}
    \bfx_{k+1} = \sigma\left(\bfx_k + \bfcalR_k\left(\bfx_k; \bftheta_k\right)\right).
\end{equation*}
The \texttt{BasicBlock} residual block is
\begin{equation*}
    \bfcalR_k\left(\bfx_k; \bftheta_k\right) := \bfcalB_2\left(\bfcalC_2\sigma\left(\bfcalB_1\left(\bfcalC_1\bfx_k\right)\right)\right),
\end{equation*}
the composition of a sequence of the following operators:
\begin{itemize}
    \item $\bfcalC$: A \textit{convolutional kernel} of size $(N_1, N_2, N_3, N_3)$ acting on an activation map $\bfx_k$ of size $(N_2, W, H)$, where $W, H > N_3$. For the residual blocks we are concerned with here, the output of $\bfcalC(\bfx_k)$ will be of size $(N_1, W, H)$. The weights of the convolutional kernels are trainable parameters. Note that convolutional kernels in ResNets do not include bias terms. 
    \item $\bfcalB$: \textit{Batch normalization}~\cite{ioffe2015batch} of the form 
    \small
    \begin{equation}
        \bfcalB(\bfz) = \frac{\bfz - \bfmu}{\bfnu}\bfgamma + \bfbeta,
    \end{equation}
    \normalsize
    where $\bfmu$ and $\bfnu$ are the batch mean and variance, respectively, and $\bfgamma$ and $\bfbeta$ are trainable parameters. All of these parameters are fixed when using the network for evaluation.
    \item $\sigma$: \textit{Nonlinearity} or \textit{activation function} that acts element-wise on its input; assumed to be ReLU since this is the activation function commonly used in conjunction with ResNets, but can be any other activation function, as long as the same activation function was used during training.
\end{itemize}
Each set of parameters $\bftheta_k$ in the $k$-th residual unit is the combination of the weights of all the convolutional kernels and parameters in the batch normalization operators in that unit.

The linearization of the update formula~\eqref{eqn:residual_unit} is
\begin{equation*}
    \bfJ = \mathrm{diag}\left(\sigma'(\bfx_k + \bfcalR(\bfx_k))\right)\left(\bfI + \bfJ_{\bfcalR}\right).
\end{equation*}
$\bfJ_{\bfcalR}$ is the linearization of the residual block:
\begin{equation*}
\bfJ_{\bfcalR} = \bfJ_{\bfcalB_2}\bfcalC_2\,\mathrm{diag}\left(\sigma'\left(\bfcalB_1\left(\bfcalC_1\,\bfx_k\right)\right)\right)\bfJ_{\bfcalB_1}\bfcalC_1.
\end{equation*}
We have let $\bfJ_{\bfcalB_i} = \frac{d\,\bfcalB_i}{d\,\bfx_k}$ denote the linearization of the $i$-th batch normalization operator, which is the same as the operator $\bfcalB_i$, but with $\bfmu$ and $\bfbeta$ set to $\bfzero$, i.e. if $\bfcalB_i(\bfz) = \frac{\bfz - \bfmu_i}{\bfnu_i}\bfgamma_i + \bfbeta_i$, then $\bfJ_{\bfcalB_i}(\bfz) = \frac{\bfgamma_i}{\bfnu_i}\bfz$.

\begin{figure}[H]
\centering
\includegraphics[width=.95\linewidth]{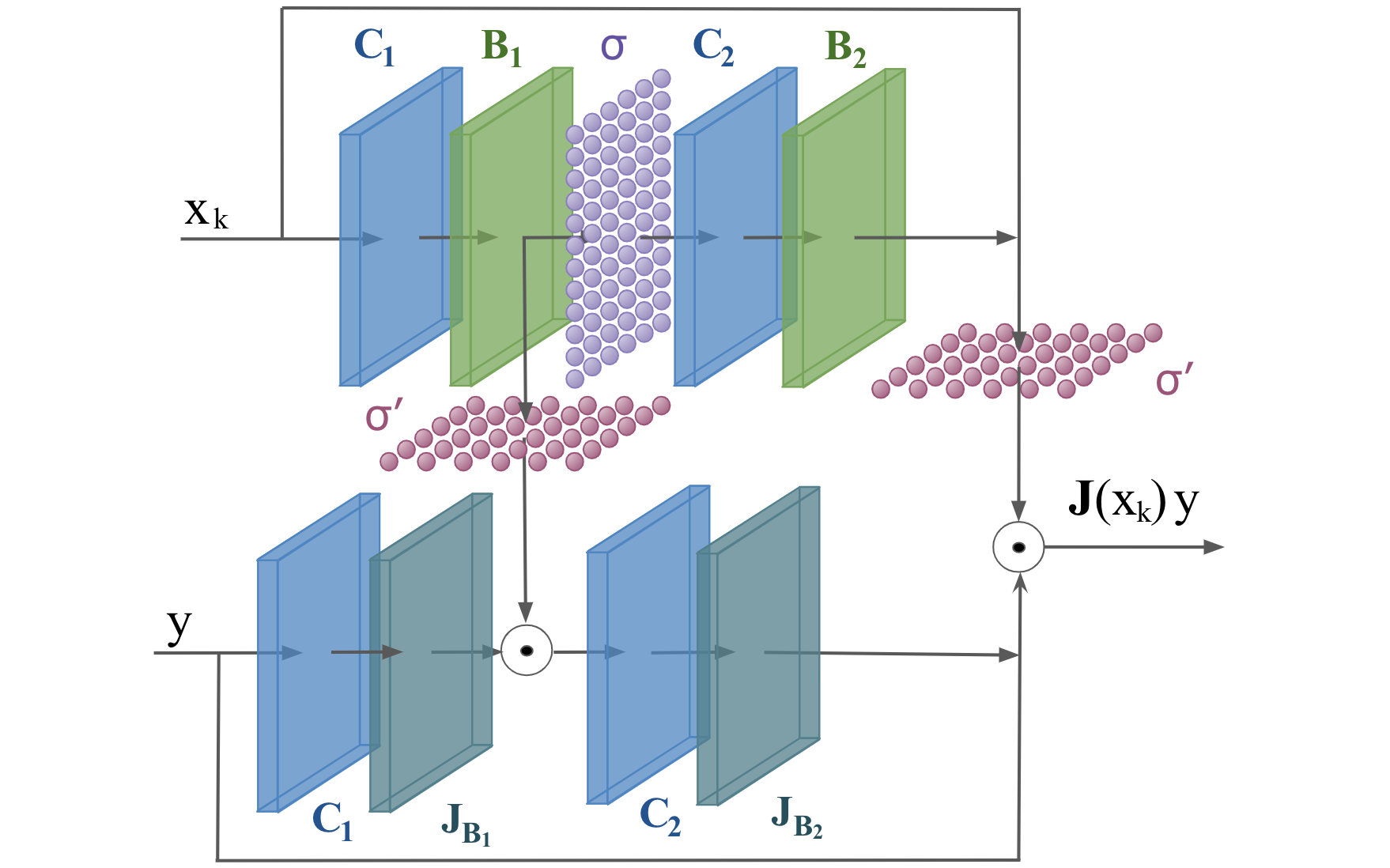}
\caption{Schematic representation of the linearization of a \texttt{BasicBlock} residual unit.}
\label{fig:basicblock_jacobian}
\end{figure}

Iterative methods require only the product of $\bfJ$ with arbitrary vectors $\bfy$ of length $N$. 
An efficient way to perform this product is to implement the Jacobian computation as its own form of residual unit that takes both the activation map $\bfx$ and the arbitrary vector $\bfy$ as input; an illustration of this computation is given in Figure~\ref{fig:basicblock_jacobian}. 
For this computation we require that $\bfy$ is reshaped to have the same dimensions as the activation map $\bfx$. 
Existing iterative methods for the singular value decomposition can then be used to determine the eigenvalues of $\bfJ$.

We note the following:
\begin{itemize}
    \item The Jacobian depends explicitly on the activation maps $\bfx$, so that its eigenvalues will depend on $\bfx$ as well. 
    This means that the system could be stable for some inputs, but unstable for others.
    \item The correctness of the implementation of the Jacobian computation should be tested using the \textit{Jacobian test}: choose an $\epsilon$ and decrease it by an order of magnitude for a small number of iterations. 
    Then the norm of the residual $\bfcalR(\bfx + \epsilon \delta \bfx) - \bfcalR(\bfx) - \epsilon\bfJ_{\bfcalR}(\bfx)\delta \bfx$ should be decreasing an order of magnitude faster than the norm of the residual $\bfcalR(\bfx + \epsilon \delta \bfx) - \bfcalR(\bfx)$.
    \item If $\sigma$ is not smooth everywhere, such as ReLU at $x = 0$, then the $\sigma'$ will have a jump discontinuity, which may manifest itself when testing for the correctness of the implementation of the Jacobian. 
    For simplicity, one can use a smooth activation function when performing the Jacobian test, but the discontinuity of ReLUs means that the residual blocks are actually not differentiable. 
    In practice this is important only for perturbations of inputs that are close to $0$ and it does not appear to matter much.
\end{itemize}
The transpose of $\bfJ$ is 
\begin{equation*}
    \bfJ^\top = \left(\bfI + \bfJ_{\bfcalR}^\top\right)\mathrm{diag}\left(\sigma'(\bfx_k + \bfcalR(\bfx_k))\right),
\end{equation*}
where 
\begin{equation*}
\bfJ_{\bfcalR}^\top = \bfcalC_1^\top\bfJ_{\bfcalB_1}^\top\,\mathrm{diag}\left(\sigma'\left(\bfcalB_1\left(\bfcalC_1\,\bfx_k\right)\right)\right)\bfcalC_2^\top\bfJ_{\bfcalB_2}^\top.
\end{equation*}
The product of $\bfJ^\top$ with arbitrary vectors, which is required by the SVD, is implemented similarly to above and requires transpose convolution operators. We omit the details here. 

\end{document}